\documentclass[10pt,twocolumn,letterpaper]{article}

\usepackage{cvpr}              

\usepackage{graphicx}
\usepackage{amsmath}
\usepackage{amssymb}
\usepackage{booktabs}

\usepackage[dvipsnames]{xcolor}
\usepackage{epsfig}
\usepackage{graphicx}
\usepackage{amsmath}
\usepackage{amssymb}
\usepackage{mathtools}
\usepackage{subcaption}
\usepackage{comment}
\usepackage{mathtools}
\usepackage{acronym}
\usepackage{caption}
\usepackage{array}
\usepackage{tabularx}
\usepackage{subcaption}
\usepackage{booktabs}
\usepackage{multicol}
\usepackage{multirow}
\usepackage{inconsolata}
\usepackage{makecell}
\usepackage[norelsize, linesnumbered, ruled, lined, boxed, commentsnumbered]{algorithm2e}

\usepackage{xspace}

\usepackage[pagebackref,breaklinks,colorlinks]{hyperref}
\definecolor{orange}{rgb}{0.6, 0.3, 0.1}
\definecolor{teal}{rgb}{0.0, 0.4, 0.4}
\definecolor{purple}{rgb}{0.65,0,0.65}
\definecolor{saffron}{rgb}{0.95,0.75,0.2}
\definecolor{turquoise}{rgb}{0.0,0.5,0.5}
\definecolor{brickred}{rgb}{.6, .2 .1}
\definecolor{coral}{rgb}{1,0.45,0.33}
\definecolor{red}{rgb}{1, 0, 0}
\definecolor{brown}{rgb}{0.8, 0.3, 0}

\makeatletter

\newcommand{\Rmnum}[1]{\expandafter\@slowromancap\romannumeral #1@}
\makeatother
\usepackage[capitalize]{cleveref}
\crefname{section}{Sec.}{Secs.}
\Crefname{section}{Section}{Sections}
\Crefname{table}{Table}{Tables}
\crefname{table}{Tab.}{Tabs.}
\crefname{figure}{Fig.}{Figs.}

\def\cvprPaperID{7042} 
\def\confName{CVPR}
\def\confYear{2024}

\begin{document}

\title{Deformable One-shot Face Stylization via DINO Semantic Guidance}

\author{Yang Zhou \qquad Zichong Chen  \qquad Hui Huang\thanks{Corresponding author.}\\
	Visual Computing Research Center, Shenzhen University\\
}




\makeatletter
\let\@oldmaketitle\@maketitle
\renewcommand{\@maketitle}{\@oldmaketitle
    \vspace{-1\baselineskip}
    \includegraphics[width=\linewidth]{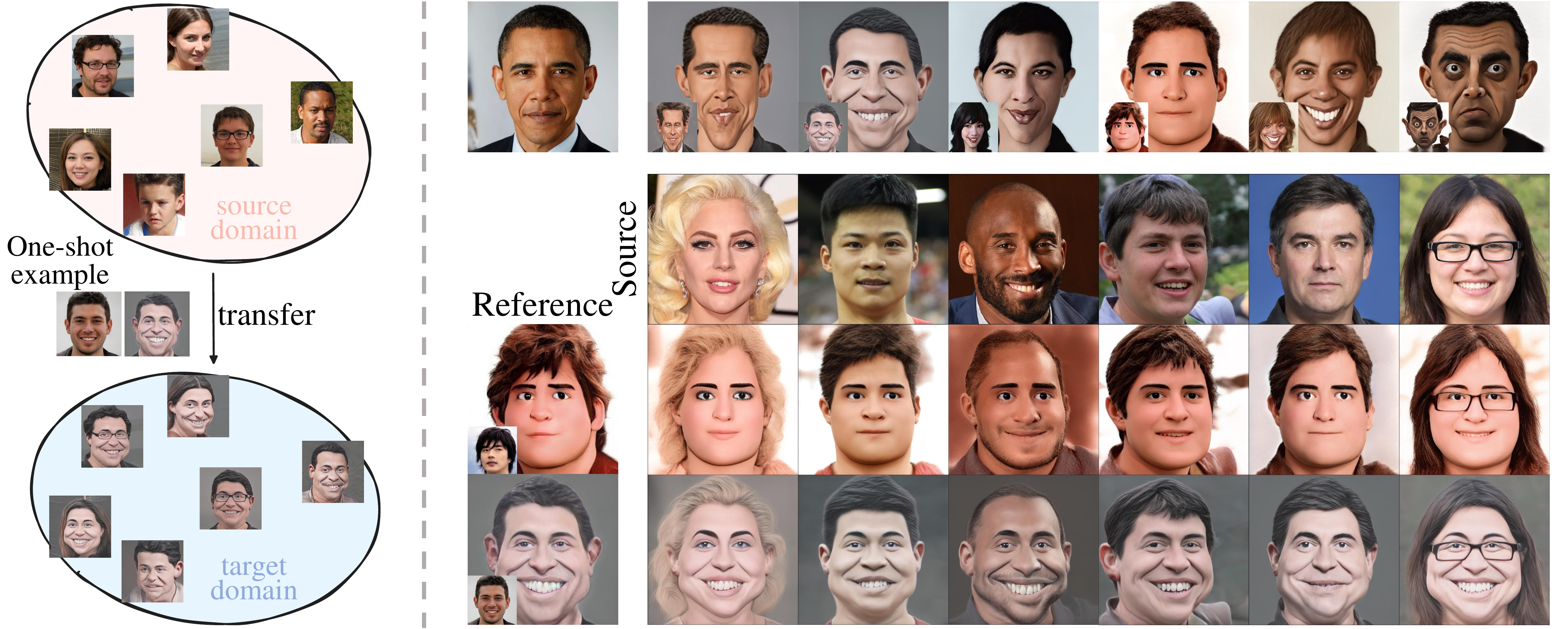}
     \captionof{figure}{We propose a \textbf{deformation-aware face stylization} framework trained on a single real-style image pair (left). Our framework can generate diverse, high-quality, stylized faces with desired deformations, while maintaining the input identity well (right).}
    \vspace{1em}
    \label{fig:teaser}
\bigskip}

\makeatother

\maketitle

\begin{abstract}

This paper addresses the complex issue of one-shot face stylization, focusing on the simultaneous consideration of appearance and structure, where previous methods have fallen short. We explore deformation-aware face stylization that diverges from traditional single-image style reference, opting for a real-style image pair instead. The cornerstone of our method is the utilization of a self-supervised vision transformer, specifically DINO-ViT, to establish a robust and consistent facial structure representation across both real and style domains. Our stylization process begins by adapting the StyleGAN generator to be deformation-aware through the integration of spatial transformers (STN). We then introduce two innovative constraints for generator fine-tuning under the guidance of DINO semantics: i) a directional deformation loss that regulates directional vectors in DINO space, and ii) a relative structural consistency constraint based on DINO token self-similarities, ensuring diverse generation. Additionally, style-mixing is employed to align the color generation with the reference, minimizing inconsistent correspondences. This framework delivers enhanced deformability for general one-shot face stylization, achieving notable efficiency with a fine-tuning duration of approximately 10 minutes. 
Extensive qualitative and quantitative comparisons demonstrate our superiority over state-of-the-art one-shot face stylization methods. Code is available at 
\href{https://github.com/zichongc/DoesFS}{https://github.com/zichongc/DoesFS}.

\end{abstract}

\vspace*{-0.5cm}

\section{Introduction}
\label{sec:intro}

Face stylization is a stylish and eye-catching application favored by users in social media and the virtual world. Recent advances~\cite{Song2021AgileGAN, yang2022Vtoonify, yang2023styleganex, yang2022Pastiche, men2022dct} mainly benefit from the capacity of generative models, \eg, StyleGAN~\cite{karras2020stylegan, karras2019stylebased, Karras2020ada}. When coping with artistic styles of extremely limited examples (\eg, one), it is necessary to prevent the training from over-fitting and mode collapse. Many one-shot methods~\cite{chong2022jojogan, zhu2022mtg, guhyun2023oneshotclip, zhang2022towards} have addressed this issue with different strategies. However, these approaches mainly focus on color and texture transfer, seldom exploring the geometric deformation potential. As exaggeration is an important characteristic of artistic style, structural deformations should also be emphasized in the stylization. 

\begin{figure}[t!]
	\centering
	\includegraphics[width=\linewidth]{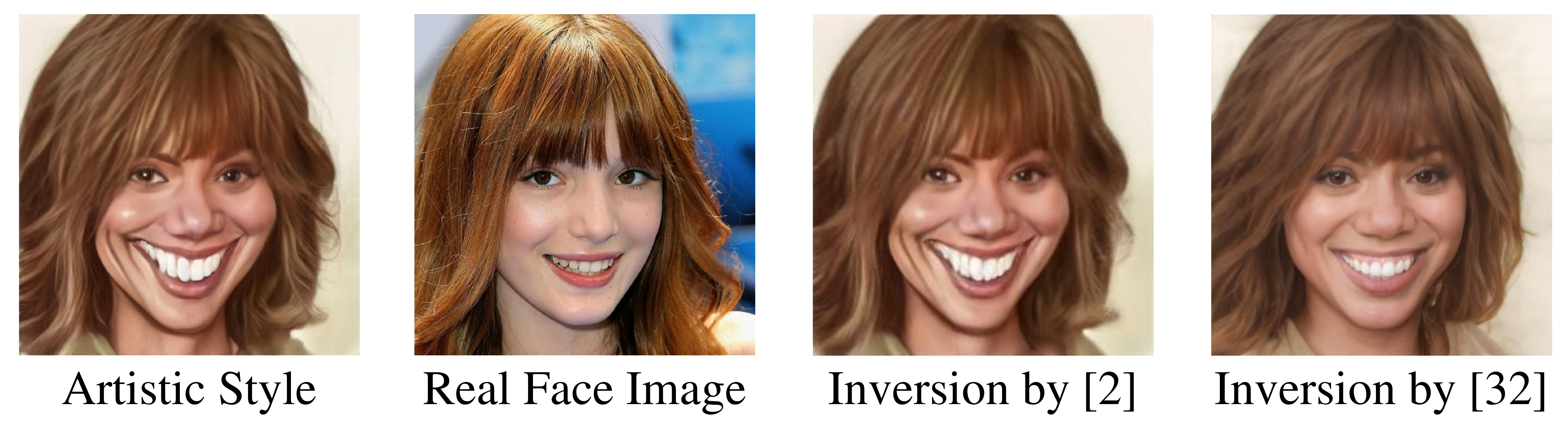}
	\caption{
        \textbf{ GAN inversion.}
		 Previous one-shot stylization methods like ~\cite{zhu2022mtg} and~\cite{chong2022jojogan}, build the cross-domain guidance by inverting the artistic style image into real face domain. But compared with the ground truth real face, current GAN inversion techniques~\cite{tov2021designing,abdal2019image2stylegan} still cannot give out a faithful mapping across unseen domains, which may mislead the structure deformation in the stylization.}
	\label{fig:gan_inversion}
    \vspace*{-2mm}
\end{figure}

Can we stylize facial images according to only one style example while simultaneously considering \emph{appearance change} and \emph{structure exaggeration}? Previous methods~\cite{chong2022jojogan, zhu2022mtg, guhyun2023oneshotclip} solve this problem by building ``fake" guidance across the source and target domain based on the inversion of the style reference. However, as shown in Fig.~\ref{fig:gan_inversion}, current GAN inversion techniques, such as~\cite{abdal2019image2stylegan} and~\cite{tov2021designing}, still can not produce faithful mappings for artistic images, which will definitely mislead the geometric deformation, especially when strong exaggerations occur. We argue that we can build reliable deformation guidance across domains using a \emph{real-style} image pair instead of only the style example, thus reducing the difficulty; see,~\eg, Fig.~\ref{fig:teaser}. Nonetheless, existing one-shot methods still failed to capture the deformation pattern because of the weaker structural guidance they employed; see Sec.~\ref{sec:experiments} for more details. 

In this work, we propose a novel stylization network trained on a single real-style image pair. The network is built upon a pre-trained StyleGAN, with additional spatial transformers appended to make the generator deformation-aware. To overcome the huge domain gap between the given paired references, we dive into the feature space of DINO~\cite{caron2021emerging}, a self-supervised vision transformer. We present evidence for DINO in capturing robust and consistent structure semantics across real and style face domains, superior to the other popular ViTs. Therefore, we optimize our generator based on the DINO semantic guidance to learn the cross-domain structural deformation. Specifically, we compute a novel directional deformation loss that regularizes the direction vector in DINO space between the real and the style faces. Meanwhile, a new relative structure consistency upon self-similarities of DINO features is calculated to ensure the diversity of the target (style) domain, preventing overfitting. Finally, color alignment based on style-mixing techniques is applied to further guarantee the correctness of semantic correspondence. Experiments show that our method can accurately stylize facial images into artistic styles with strong exaggerations, both in appearance change and structure deformation, yet still maintaining a faithful identity to the input.

The main contributions are summarized as follows:
\begin{itemize}
    \vspace{-2mm}
    \item We explore the feature space of DINO and discover its powerful structural/semantic representation both in real and style face domains.
    \vspace{-2mm}
    \item Based on DINO features, we propose two novel cross-domain losses to constrain the geometric deformation from real faces to artistic styles.
   \vspace{-2mm}
    \item We propose a novel deformable face stylization network, trained with only a single-paired real-style example. Extensive qualitative and quantitative comparisons with existing state-of-the-art demonstrate the effectiveness and superiority of our framework.
\end{itemize}

\section{Related Work}

\paragraph{Face Stylization.} Driven by deep learning techniques, face stylization has evolved rapidly. Since the seminal work of~\cite{gatys2016ImageStyleTransfer}, methods upon neural style transfer have been developed for face stylization~\cite{huang2017adain, Kim2020DST, song2019ETNet, wu2020EFNet, kolkin2019strotss}. However, these general style transfer methods did not make use of the learned prior of generative models, \eg, GANs~\cite{goodfellow2014gan}. In contrast, image translation-based methods~\cite{shi2019warpgan, Jang2021StyleCari, yang2022Pastiche, pinkney2020resolution, Song2021AgileGAN, cao2018cari, Gong2020AutoToonAG, men2022dct} usually train dedicated stylization GANs, or fine-tune pretrained StyleGANs~\cite{karras2020stylegan, karras2019stylebased} over large collections of artistic facial images (from hundreds to thousands). Although some of these methods are deformation-aware, specifically for caricature portraits, the requirement of data amount makes them fail in styles defined by extremely limited examples, which is yet the most cases in real applications. Recently, researchers have addressed the few data challenge on domain adaption by introducing a series of regularizations in the fine-tuning~\cite{ojha2021cdc, Xiao2022rssa, Zhao2022DCL, zhao2022adam, zhao2023exploring, mondal2023fewshot}. On face stylization, more specifically, a few works~\cite{chong2022jojogan, zhu2022mtg, guhyun2023oneshotclip, zhang2022towards} further use a single example as the style reference. 
Although they can generate well-stylized appearances, the one-shot methods tend to strictly preserve the face structures, while overlooking the non-local deformations in the style example. In this work, we present a novel framework for deformable face stylization based on a single paired data. 

\paragraph{ViT Feature Representation.} 
Features from vision transformer (ViT)~\cite{dosovitskiy2020vit} are powerful and versatile visual representations. Researchers have demonstrated that ViT trained in specific manners can serve well for numerous downstream tasks in both vision and graphics~\cite{radford2021clip, zheng2022farl, caron2021emerging, amir2021deep}. Among them, CLIP~\cite{radford2021clip} are most widely used for text-guided image editing~\cite{patashnik2021styleclip}, generation~\cite{rinon2022stylegannada}, and stylization~\cite{zhu2022mtg, zhang2022towards, guhyun2023oneshotclip }. Then, DINO~\cite{caron2021emerging}, a self-supervised ViT model, exhibits striking properties in capturing high-level semantic information~\cite{amir2021deep}. Amir~\etal~\cite{amir2021deep} use the keys of DINO as ViT features and apply them to many challenging vision tasks in unconstrained settings. Tumanyan~\etal~\cite{tumanyan2022splicing} further splice the DINO features as disentangled appearance and structure representations, realizing a semantic-aware appearance transfer from one natural image to another. Given the success achieved, we believe DINO has the potential for deformable face stylization. Next, we will explore the DINO features and compare them with two other ViT features.

\section{DINO Semantic Guidance}
\label{sec:dino}

In this section, we first briefly review the DINO-ViT~\cite{caron2021emerging}. Then, we explore and analyze the properties of different DINO features, and further compare them with two weakly supervised ViTs, CLIP~\cite{radford2021clip} and FaRL~\cite{zheng2022farl}.

DINO is a ViT model trained in self-distillation~\cite{caron2021emerging}. During training, an input image is randomly transformed into two different inputs for a student and a teacher network (with the same architecture but different parameters). The student is optimized by cross-entropy loss that measures the output similarity, and the exponential moving average of the student updates the teacher. DINO shows emerging properties in encoding high-level semantic information at high spatial resolution~\cite{caron2021emerging, amir2021deep}. We expect DINO to extract a consistent semantic representation across the source and target domains (\ie, the real and the style faces). Thus, we can build reliable deformation guidance despite the huge domain gap.

Previously, Amir~\etal~\cite{amir2021deep} demonstrated that DINO is superior to traditional CNN-based models and supervised ViTs in semantic information capture. But they lack the comparison with alternative weakly-supervised ViTs, such as CLIP~\cite{radford2021clip}, which has been widely adopted in many style transfer tasks~\cite{patashnik2021styleclip,rinon2022stylegannada}. To investigate comprehensively, we choose two recently popular weakly-supervised ViTs, CLIP and FaRL~\cite{zheng2022farl}, which are both text-guided and learn a multi-modal embedding to estimate the semantic similarity between a given text and an image. In particular, FaRL is specially targeted for human facial representation.

\begin{figure}[t!]
    \centering
    \includegraphics[width=\linewidth]{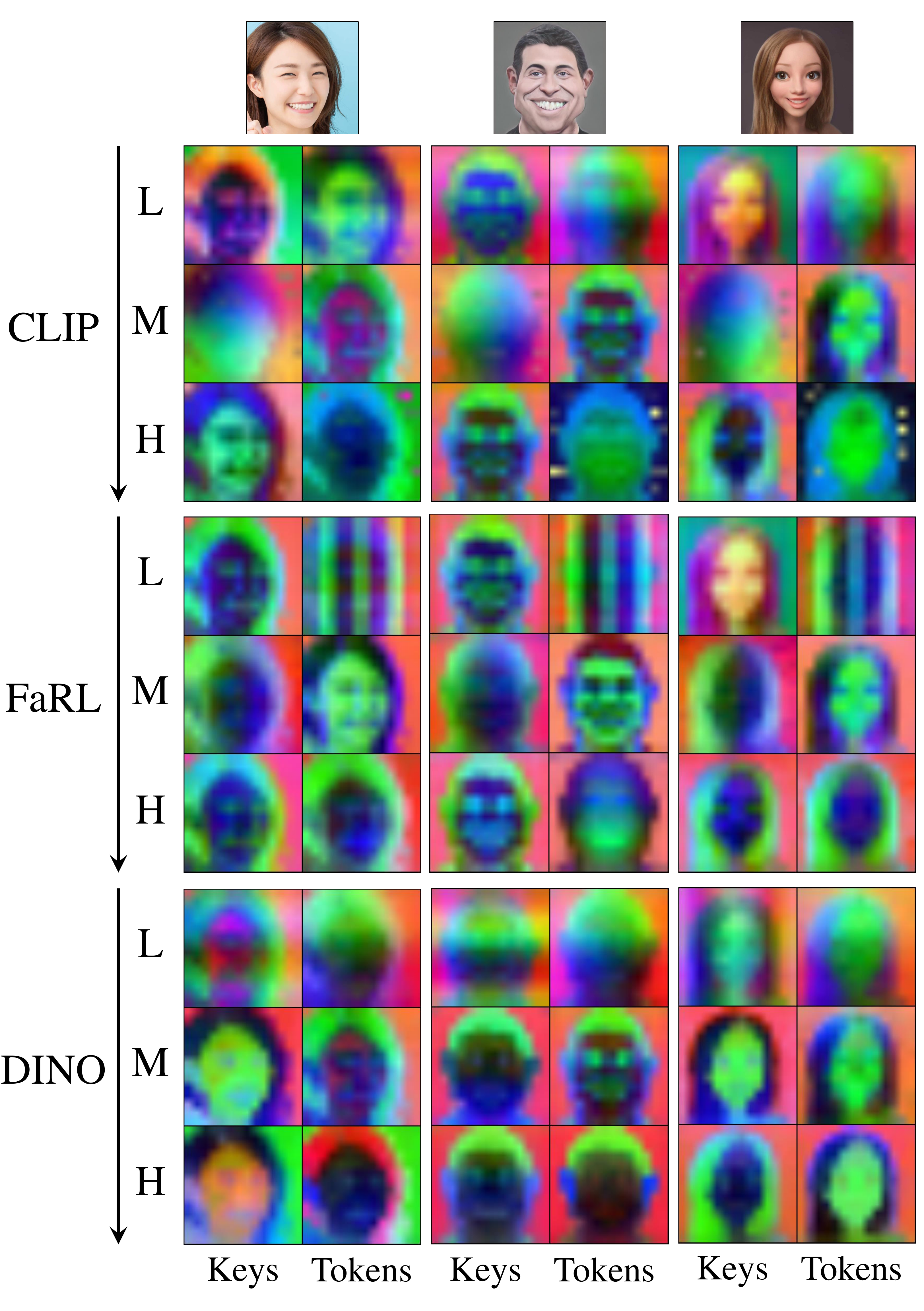}
    \vspace*{-7mm}
    \caption{
    \textbf{Visualization of the hierarchical features} from CLIP~\cite{radford2021clip}, FaRL~\cite{zheng2022farl}, and DINO~\cite{caron2021emerging}, where the same color represents the same semantics shared. Note that the same architecture ViT (ViT-B/16) is employed fairly for image encoding. We choose layers 3, 6, and 12 to represent different levels (L, M, H) of features. Following~\cite{amir2021deep}, we only use keys and tokens in ViTs, while discarding the queries and values. The \texttt{[CLS]} token is also discarded as it mainly encodes the visual appearance~\cite{tumanyan2022splicing}. 
    }
    \label{fig:pca}
    \vspace*{-5mm}
\end{figure}

So, are DINO features better than those of existing weakly-supervised ViTs (\ie, CLIP and FaRL) for semantic representation in the task of face stylization? 
To this end, we visualize their intermediate features for comparison. As shown in Fig.~\ref{fig:pca}, we input both real and artistic face images to these ViTs. Keys and tokens from different layers are visualized by principal component analysis (PCA). From the PCA results, high-level DINO features show cleaner, more precise, and, more importantly, more consistent facial segments than CLIP and FaRL features. Lower-level features of DINO contain semantic and positional information simultaneously. In contrast, features from CLIP and FaRL are relatively chaotic, demonstrating the superiority of DINO.

We conclude that DINO can well capture the facial semantics across different facial domains, whereas others cannot. We attribute the advantages to DINO's self-distillation. The pair-wised augmentation on the training input makes DINO naturally focus on structural semantics. On the contrary, text-guided ViTs lose precision in semantic capture, probably due to the ambiguity of natural language. Furthermore, different from~\cite{tumanyan2022splicing, amir2021deep}, we found using keys will also miss some structural semantics through a simple over-fitting test; see supplementary for more details and a comprehensive comparison. Therefore, we use DINO tokens as our feature representation, named DINO semantic guidance.

\section{Method}
\label{sec:method}

\begin{figure*}[htbp]
    \centering
    \includegraphics[width=1.\linewidth]{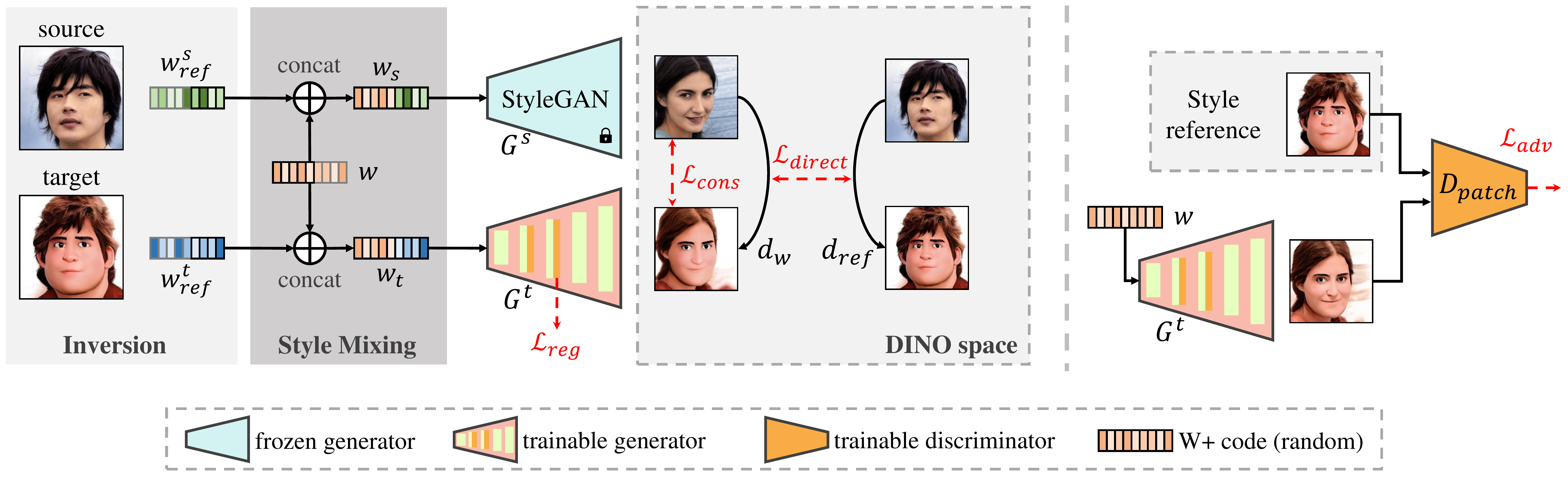}
        \caption{
    \textbf{ Framework Overview.} Given a single real-style paired reference, we fine-tune a deformation-aware generator $G^t$ that simultaneously realizes geometry deformation and appearance transfer. To learn the cross-domain deformation, we design a directional deformation loss $\mathcal{L}_{direct}$ and a relative structural consistency loss $\mathcal{L}_{cons}$, both computed in DINO feature space (middle). Inversion and style mixing further ensure a consistent DINO semantic representation aligned with the given reference (left). In addition, we involve adversarial training using a patch-level discriminator to enhance the transferred style and fidelity (right). } 
    \label{fig:pipeline}
    \vspace*{-2mm}
\end{figure*}

As illustrated in Fig.~\ref{fig:pipeline}, at the core of our framework is a deformation-aware generator $G^{t}$ fine-tuned by two novel DINO-based losses and an adversarial loss. Specifically, we design $G^{t}$ upon StyleGANv2~\cite{karras2020stylegan}, with spatial transformers~\cite{max2015stn} appended for structural deformation. During training, we first sample a latent code $w\in \mathbb{R}^{18\times 512}$ in W+ space, followed by style mixing for color alignment. Then, we feed the code to $G^{t}$ and $G^{s}$, respectively, where $G^{s}$ is a pre-trained StyleGAN generator frozen during training, and $G^{t}$ is to be fine-tuned by the new proposed losses. In inference time, we first input the given facial images to a pre-trained e4e encoder~\cite{tov2021designing}, obtaining their inversion code $w^*$ in W+ space. Then, we can generate the corresponding stylized faces $G^t(w^*)$ by a single forward pass.


Below, we first describe the details of our deformation-aware generator, followed by the DINO-based domain adaption, where two novel constraints based on DINO semantic guidance are introduced. In the end, we describe the color alignment enhancing the stability. 

\begin{figure}[t]
    \centering
    \includegraphics[width=1.\linewidth]{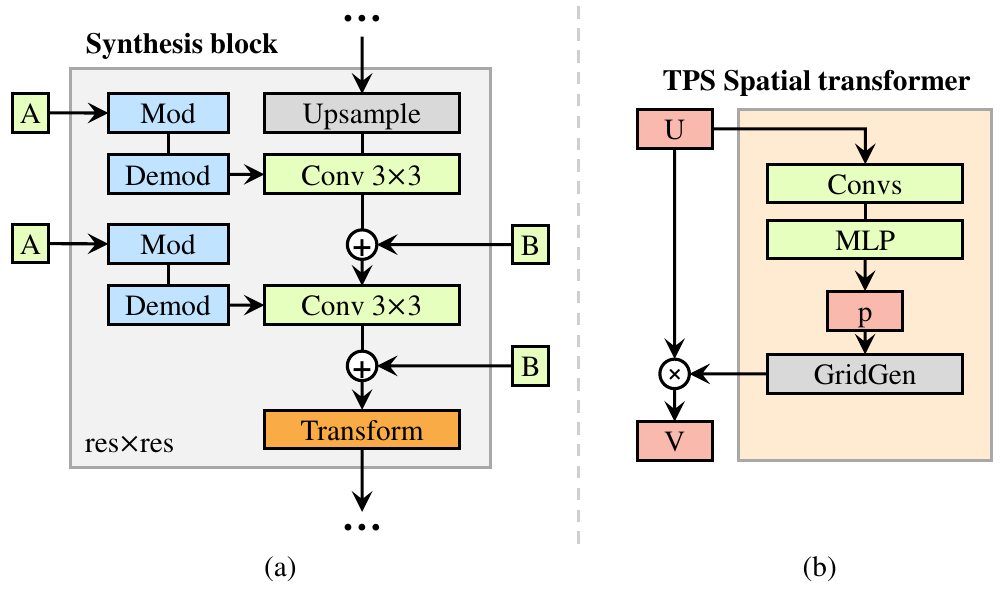}
        \caption{
    \textbf{ Deformation-aware generator}. (a) A StyleGAN synthesis block plugged in with a \texttt{Transform} module. (b) A thin-plate-spline spatial transformer (TPS-STN) for feature warping, which shows the detail of the \texttt{Transform} module in (a).}
    \label{fig:styleganstn}
    \vspace*{-2mm}
\end{figure}

\subsection{Deformation-aware generator}
\label{sec:generator}
StyleGAN~\cite{karras2020stylegan} pretrained on large-scale facial datasets, such as FFHQ~\cite{karras2019stylebased}), is powerful in various facial image generation tasks. However, it is hard to synthesize faces with exaggerated components given the prior learned from real face domain. Inspired by~\cite{yang2022Vtoonify}, we warp the intermediate features of StyleGAN to facilitate the generator to output the desired deformation pattern.

As depicted in Fig.~\ref{fig:styleganstn}, we insert a simple differentiable spatial transformer network (STN)~\cite{max2015stn}, denoted as \texttt{Transform}, after the synthesis block of StyleGAN generator. The \texttt{Transform} module transforms a feature map by a single forward pass, where the transformation could be translation, rotation, thin-plane-spline warping (TPS), etc. In our case, we chose TPS-STN for warping feature maps, as well as a basic STN that shares the same architecture as TPS-STN to perform translation, rotation, and scaling. Empirically, we only add \texttt{Transform} in the resolutions of $32\times32$ and $64\times64$ with a grid size of $10$.

We are not the first to make StyleGAN deformation-aware. StyleCariGAN~\cite{Jang2021StyleCari} presented a layer-mixed StyleGAN with CNN-based exaggeration blocks. However, these deformation blocks require additional training with massive training data. Our STN-based \texttt{Transform}, instead, only performs simple transformations on feature maps, which can be directly plugged into the pretrained StyleGAN generator. According to \cite{Gong2020AutoToonAG}, we regularize the TPS warping field to be smooth by
\begin{equation}
    \mathcal{L}_{reg}=\sum_{i,j\in \textbf{F}}\Big(2-{\rm sim}(\textbf{F}_{i,j-1}, \textbf{F}_{i, j})-{\rm sim}(\textbf{F}_{i-1,j},\textbf{F}_{i,j})\Big),
 \end{equation}
where ${\rm sim}$ refers to the cosine similarity, $\textbf{F}$ represents the warping field, and $i,j$ are pixel indices.

\subsection{DINO-based domain adaption} 
\label{sec:adaption}
Our deformation-aware generator is initialized by a StyleGANv2 model pre-trained on FFHQ. To adapt it to the target domain defined by the style reference, we fine-tune the generator with three criteria: 1) a novel directional deformation loss guiding the cross-domain structural deformation, 2) a newly designed relative structural consistency transferring the generation diversity, and 3) an adversarial loss facilitating the style synthesis.

\paragraph{Directional Deformation Loss.}
Based on the DINO semantic guidance in Sec.~\ref{sec:dino}, we construct a directional deformation loss guiding the cross-domain structural deformation. As in Fig.~\ref{fig:pipeline}, given a single paired reference, we first project them into DINO space and calculate the structural changing direction from source to target, represented by $\textbf{d}_{ref}=E_D(I_{ref}^t)-E_D(I_{ref}^s)$, as the directional deformation reference. We expect the deformation direction of generated images across domains to align with the deformation reference. Akin to \cite{rinon2022stylegannada}, we compute the DINO-based directional loss measured by cosine similarity between the cross-domain deformation directions, which is 
\begin{equation}
    \mathcal{L}_{direct}=1-{\rm sim}(\frac{\textbf{d}_w}{||\textbf{d}_w||},\frac{\textbf{d}_{ref}}{||\textbf{d}_{ref}||}),
\end{equation}
where $\textbf{d}_w=E_D(G^t(w))-E_D(G^s(w))$, with $E_D$ as the DINO encoder, and $w$ is the latent code in W+ space. 

\begin{figure}[t]
    \centering
    \includegraphics[width=1.\linewidth]{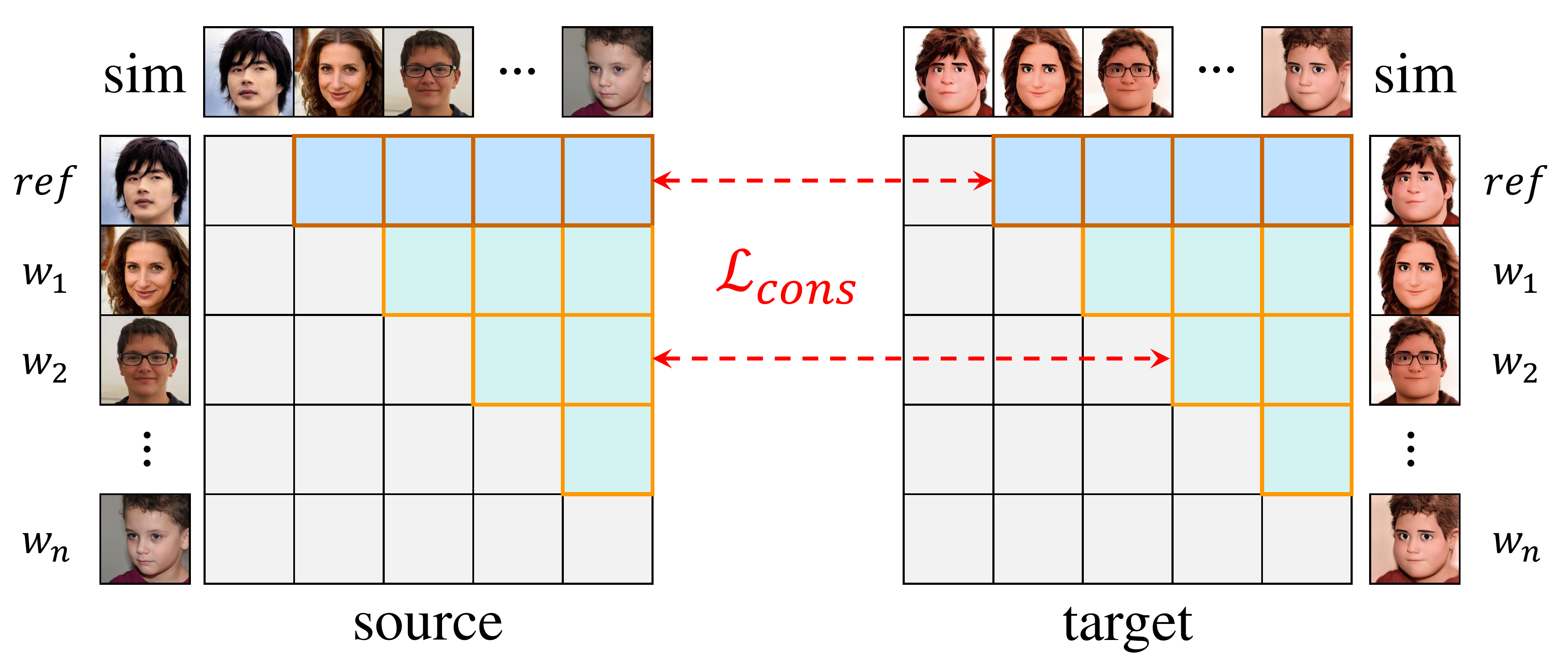}
        \caption{
    \textbf{ Relative structural consistency. } We compute a relative structural similarity matrix for each domain, where pairs of generated-reference (in blue) and generated-generated (in green) are both considered. A well-designed loss $\mathcal{L}_{cons}$ is dedicated to preserving similar structural diversity between the two domains. }
    \label{fig:consistency}
    \vspace*{-2mm}
\end{figure}

\paragraph{Relative Structural Consistency.}
As a few-shot training, adapting the generator only with the above directional deformation guidance will easily result in over-fitting. Inspired by \cite{ojha2021cdc}, we introduce a relative structural consistency to preserve the structural diversity across domains. 
Specifically, we measure the difference/similarity between every two generated batch samples of the same domain. The self-similarity vector\footnote{The self-similarity between DINO tokens forms a large similarity matrix (in size of 783*783). We flatten it into a self-similarity vector.} computed with DINO tokens is used as the structure representation for each sample, which is $s_i=\frac{E_D(I_i)E_D(I_i)^T}{||E_D(I_i)||^2_2}$. For the ground truth pair, we involve them in the relative structural consistency as well. Accordingly, two relative similarity matrices can be formed for the source and target domain, respectively, as depicted in Fig.~\ref{fig:consistency}. We then transform them into probability distributions by
\begin{equation}
\begin{aligned}
C^s=&{\rm softmax}\big(\big\{{\rm sim}(s_i^s, s_j^s)\big\}\big)_{{\forall i > j, 
    \atop
    i,j=1,...,N}}\\
    &\cup {\rm softmax}\big(\big\{{\rm sim}(s_i^s, s_{ref}^s)\big\}\big)_{{\forall i=1,...,N}},\\
C^t=&{\rm softmax}\big(\big\{{\rm sim}(s_i^t, s_j^t)\big\}\big)_{{\forall i > j, 
    \atop
    i,j=1,...,N}}\\
    &\cup {\rm softmax}\big(\big\{{\rm sim}(s_i^t, s_{ref}^t)\big\}\big)_{{\forall i=1,...,N}},
\end{aligned}
\end{equation}
where ${\rm sim}$ is cosine similarity, $N$ is the batch-size, superscripts $s$ and $t$ indicate the samples generated from $G^s$ and $G^t$, respectively. To encourage the generated images of target domain to have a similar structural diversity as the source, we compute the MSE loss between them, which is 
\begin{equation}
    \mathcal{L}_{cons}=\frac{1}{|C|}\sum_{i=1}^M||C_i^t-C_i^s||^2_2.
\end{equation}


\paragraph{Adversarial Style Transfer.}
Given the DINO features employed, the above two components mainly correspond to structural deformation. To guarantee a correct color transfer for face stylization, we introduce a patch-level discriminator $D_{patch}$ according to~\cite{ojha2021cdc}. Unlike image-level discriminators, the patch-level discriminator focuses more on local texture and color. Meanwhile, we found that a discriminator also helps to improve the fidelity of the generated stylized faces, thanks to the adversarial training. Following~\cite{ojha2021cdc}, we finetune a StyleGANv2 discriminator pretrained on FFHQ, and read off the layers with effective patch size of 22$\times$22. The adversarial losses for $D_{patch}$ and our deformation-aware generator $G^t$ are given by:
\begin{equation}
\begin{aligned}
    \mathcal{L}_{adv}^{D}=\log(1-D_{patch}(I_{ref}^t))+\mathbb{E}_{w}[\log(D_{patch}&(G^t(w)))],\\
    \mathcal{L}_{adv}^{G}=-\mathbb{E}_{w}[\log(D_{patch}(G^t(w)))].
\end{aligned}
\end{equation}

The total loss function can be finalized as:
\begin{equation}
\mathcal{L}_{total}=\mathcal{L}_{adv}+\lambda_{direct}\mathcal{L}_{direct}+\lambda_{cons}\mathcal{L}_{cons}+\lambda_{reg}\mathcal{L}_{reg}.
\end{equation}


\subsection{Color alignment}
\label{sec:alignment}
Even though DINO well disentangles the structure and appearance of images in its feature space, the vast color variations in the real facial domain would bring undesired errors in structure matching. To alleviate this disturbance, we use the inversion and style mixing technique of StyleGAN to align the samples' color. 
As shown in the leftmost of Fig.~\ref{fig:pipeline}, we first inverse the paired reference $(I_{ref}^s, I_{ref}^t)$ into the latent space of source domain by optimizing from a mean latent $\bar{w}$ using L1 and perceptual loss~\cite{zhang2018perceptual}, obtaining their latents $w_{ref}^s, w_{ref}^t \in \mathbb{R}^{18\times 512}$. Then, for a random sample $w$ in W+ space, unlike~\cite{chong2022jojogan, zhu2022mtg, guhyun2023oneshotclip}, we swap the fine-level (9-18th) codes of $w$ with the corresponding codes of $w_{ref}^s$ and $w_{ref}^t$, resulting in the latent codes $w_{s}, w_{t}$ with aligned color to the references. The generated images after color alignment share the same structure/identity as the original random sample but have aligning colors to the corresponding reference. See supplementary for more details.



\section{Experiments}
\label{sec:experiments}

\paragraph{Implementation Details.} Our framework is built upon the StyleGANv2~\cite{karras2020stylegan} and initialized by the model pretrained on FFHQ~\cite{karras2019stylebased}. We use $\lambda_{direct}=6$, $\lambda_{cons}=5e4$ and $\lambda_{reg}=1e$-$6$, empirically. For training, we set the batch size to 4 and used ADAM optimizer with a learning rate $0.002$ for the generator following the existing one-shot methods~\cite{chong2022jojogan, zhu2022mtg, zhang2022towards, guhyun2023oneshotclip}. For the multi-resolution STN modules, we set the learning rate of TPS-STN and basic-STN to $5e$-$6$ and $1e$-$4$, respectively, which are different from the generator, but they are fine-tuned together. All the STNs appended consist of two convolution layers followed by two linear layers. 
We use a mixture of M- and H-level features of DINO for the computation of the directional deformation guidance and only M-level features for the relative structural consistency, as we expect to keep more positional information while changing the semantic contents. All experiments are performed using a single NVIDIA RTX 3090.

For simplicity, the paired references we used are mainly obtained from existing generative models that are trained with massive training data, such as~\cite{Jang2021StyleCari, Song2021AgileGAN, Abdal20233DAvatarGANBD, yang2022Pastiche}. These models can stylize facial images with convincing quality. Their produced results are enough to represent real pairs.

\begin{figure}[t]
    \centering
    \includegraphics[width=1.\linewidth]{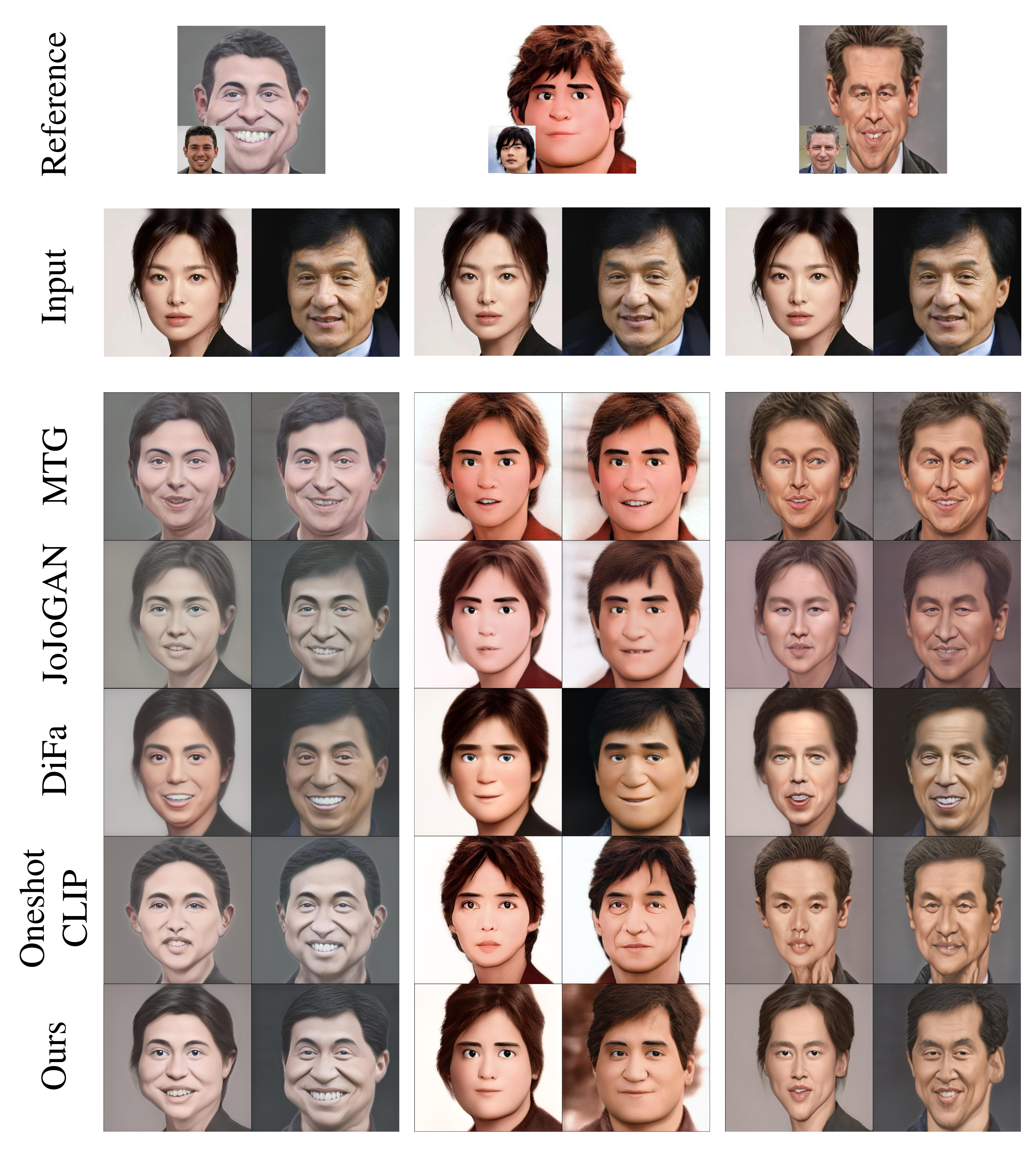}
        \caption{
     \textbf{Qualitative comparison with existing one-shot face stylization methods}, MTG~\cite{zhu2022mtg}, JoJoGAN~\cite{chong2022jojogan}, DiFA~\cite{zhang2022towards} and Oneshot-CLIP~\cite{guhyun2023oneshotclip}. Note all these competitors were trained using only the style example, whereas we used paired reference.
    }
    \label{fig:cmp_unpair}
    \vspace*{-2mm}
\end{figure}

\begin{figure*}[t]
    \centering
    \includegraphics[width=1.\linewidth]{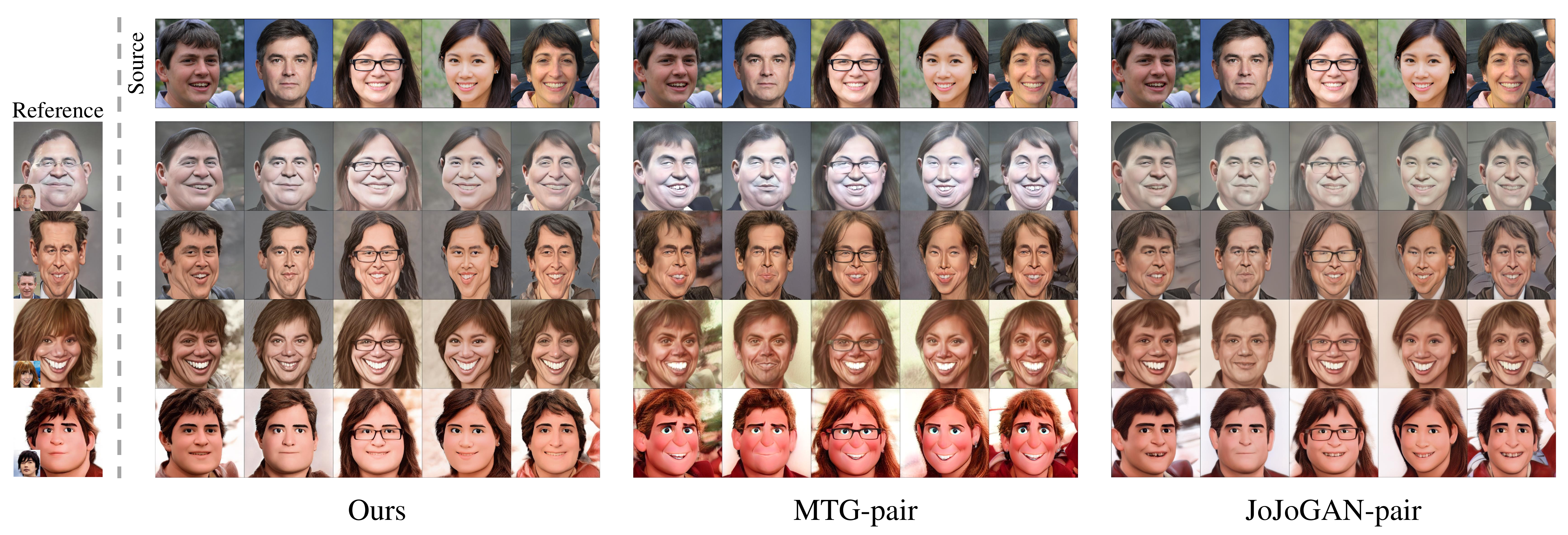}
        \caption{
     \textbf{Qualitative comparison with the ``paired'' variants} of MTG~\cite{zhu2022mtg} and JoJoGAN~\cite{chong2022jojogan}. We replace the inversion of the style example they used with the ground-truth real face. Our method surpasses these two variants both in color style and geometry deformation.
    }
    \label{fig:cmp_pair}
\end{figure*}

\begin{table*}[htbp]%
\small 
\caption{
\textbf{Quantitative comparison.} We choose the three styles shown in Fig.~\ref{fig:cmp_unpair} for evaluation, denoted as Example-1, Example-2 and Example-3. Part I shows the comparison with existing one-shot methods trained with only the style reference, whereas Part II lists the comparison with the ``paired'' variants of MTG~\cite{zhu2022mtg} and JoJoGAN~\cite{chong2022jojogan}. Note the best results are \textbf{blod}, and the second-bests are \underline{underlined}. 
} 
\label{tab:eval}
\vspace*{-5mm}
\begin{center}
\begin{tabular}{c|c|ccc|ccc|ccc}

  \Xhline{1pt}
  & \multirow{2}*{Method} & \multicolumn{3}{c|}{Example-1} & \multicolumn{3}{c|}{Example-2} & \multicolumn{3}{c}{Example-3} \\ \cline{3-11}
  & ~ & LPIPS $\downarrow$ & \makecell{dir-CC $\uparrow$} & \makecell{dir-ID $\uparrow$} & LPIPS $\downarrow$ & \makecell{dir-CC $\uparrow$} & \makecell{dir-ID $\uparrow$} & LPIPS $\downarrow$ & \makecell{dir-CC $\uparrow$} & \makecell{dir-ID $\uparrow$} \\ \Xhline{0.6pt}
  \multirow{5}*{\uppercase\expandafter{\romannumeral1}} & MTG \cite{zhu2022mtg} & \textbf{0.340} & 0.123 & 0.264 & \underline{0.361} & 0.111 & 0.301 &\textbf{0.328} & 0.095 & 0.254 \\
  & JoJoGAN \cite{chong2022jojogan} & 0.357 & 0.149 & 0.290 & 0.385 & \underline{0.138} & \underline{0.378} & 0.355 & 0.097 & \textbf{0.382} \\
  & DiFa \cite{zhang2022towards} & 0.362 & 0.142 & 0.223 & \textbf{0.293} & 0.137 & 0.328 & 0.346 & 0.094 & 0.123 \\
  & OneshotCLIP \cite{guhyun2023oneshotclip} & 0.452 & \underline{0.157} & \underline{0.333} & 0.449 & 0.119 & 0.113 & 0.460 & \underline{0.136} & 0.201 \\ \cline{2-11}
  & Ours & \underline{0.353} & \textbf{0.196} & \textbf{0.468} & 0.379 & \textbf{0.194} & \textbf{0.379} & \underline{0.342} & \textbf{0.191} & \underline{0.330} \\
  
  \hline\hline

  \multirow{3}*{\uppercase\expandafter{\romannumeral2}} & MTG-pair & 0.423 & \underline{0.195} & 0.392 & 0.447 & \underline{0.174} & 0.335 & \underline{0.379} & 0.159 & 0.280 \\
   & JoJoGAN-pair & \underline{0.381} & 0.180 & \underline{0.405} & \underline{0.395} & 0.155 & \textbf{0.393} & 0.383 & \underline{0.173} 
 & \textbf{0.400} \\ \cline{2-11}
  & Ours & \textbf{0.353} & \textbf{0.196} & \textbf{0.468} & \textbf{0.379} & \textbf{0.194} & \underline{0.379} & \textbf{0.342} & \textbf{0.191} & \underline{0.330} \\
  \Xhline{1pt}
\end{tabular} 
\end{center}
\vspace*{-3mm}
\end{table*}%


\begin{figure*}[htbp]
    \centering
    \includegraphics[width=1.\linewidth]{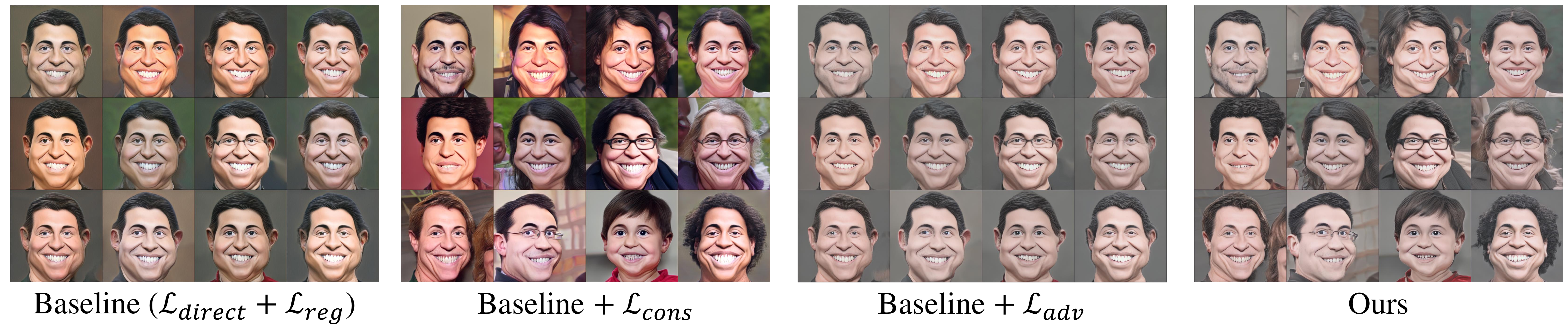}
        \caption{
    \textbf{ Ablation of different loss terms.} Using only $\mathcal{L}_{direct}$+$\mathcal{L}_{reg}$ makes the generated results all look similar to the reference. Adding $\mathcal{L}_{cons}$ improves the diversity but still lacks a correct color style regarding the reference, which is rectified after $\mathcal{L}_{adv}$ is introduced.
    }
    \vspace*{-2mm}
    \label{fig:ablation}
\end{figure*}

\subsection{Evaluation and comparisons}
\label{sec:performance}

To validate the effectiveness of our framework, we conducted qualitative and quantitative comparisons to several baselines, including MTG~\cite{zhu2022mtg}, JoJoGAN~\cite{chong2022jojogan}, DiFa~\cite{zhang2022towards}, and OneshotCLIP~\cite{guhyun2023oneshotclip}. For fairness, we also convert MTG and JoJoGAN into paired one-shot methods for comparison.

\paragraph{Qualitative Comparison.}
Fig.~\ref{fig:cmp_unpair} shows the results of stylizing two real portraits into three different styles using different adaption methods. Due to the unfaithful inversion of style reference, the compared one-shot methods fail to capture the exaggerated components, \eg, the face contour in all three examples. In contrast, our method stylizes the faces with plausible exaggeration and correct color transfer. 

We also convert MTG and JoJoGAN to accept paired references in training; see supplementary for the details of these two variants (MTG-pair and JoJoGAN-pair). Fig.~\ref{fig:cmp_pair} shows the qualitative comparison. Again, our results have better color style and finer geometry deformation. The two variants are still limited to generating deformed faces due to the lack of precise cross-domain structure guidance. More comparisons are included in the supplementary.

\paragraph{Quantitative Comparison.} 
We evaluate the generated results from three aspects: perception, deformation, and identity. For perceptual evaluation, we compute the widely used LPIPS distance~\cite{zhang2018perceptual}. To evaluate structural deformation and identity, we designed two new metrics: directional content consistency (dir-CC) and directional identity similarity (dir-ID). For both metrics, we first compute a directional feature vector for each pair between the source and the target domain. As the reference pair provides the ground truth directional vector, we measure the cosine similarity between the ground truth pair and the generated pairs. Given that we have already used DINO features in our training, for fairness, we employ VGG~\cite{Simonyan2015vgg} and ArcFace~\cite{deng2019arcface} features for these two metrics.

Table~\ref{tab:eval} lists the full quantitative evaluation. Specifically, our method shows an obvious advantage in deformation and identity compared with the one-shot methods trained with only the style reference. While compared with the ``paired'' variants, our method still leads ahead on most metrics, especially style perception and structure deformation.


\paragraph{User Study.}
We further conducted a user study to investigate human evaluation of the results. Each time, we show the user the paired reference, three real face images, three results from our method, and three results from a competitor. The user is asked to choose the better results group or ``comparable''. Table~\ref{tab:user} reports the user preference score of our method against the opponents.

\begin{table}
\small
\caption{
\textbf{User preference score. } Each user answered 60 questions, and each question asked to compare our method to an opponent on 3 real faces. Finally, 30 participants were involved, resulting in 1800$\times$3 comparisons. See supplementary for more details.
}
\label{tab:user}
\vspace*{-5mm}
\begin{center}
\begin{tabular}{c|ccc}
  \Xhline{1pt}
  Ours vs. & MTG~\cite{zhu2022mtg} & JoJoGAN~\cite{chong2022jojogan} & OneshotCLIP~\cite{guhyun2023oneshotclip} \\ \hline
  Rate & 73.0\% & 77.0\% & 83.7\%   \\ \hline\hline
  Ours vs. & MTG-pair & JoJoGAN-pair & DiFa~\cite{zhang2022towards} \\ \hline
  Rate & 76.7\% & 70.7\% & 71.3\%   \\

  \Xhline{1pt}
\end{tabular} 
\end{center}
\vspace*{-5mm}
\end{table}

\subsection{Ablation Analysis}
\label{sec:ablation}

\noindent\textbf{Effect of loss terms.}
As shown in Fig.~\ref{fig:ablation}, the directional deformation loss $\mathcal{L}_{direct}$ brings reliable deformation guidance but, meanwhile, leads to over-fitting. The relative structural consistency $\mathcal{L}_{cons}$ ensures a correct correspondence across domains but introduces color variations too. The adversarial loss stabilizes the style quality. Therefore, our full loss terms guarantee the deformation and style simultaneously.

\begin{figure*}[t]
    \centering
    \includegraphics[width=1.\linewidth]{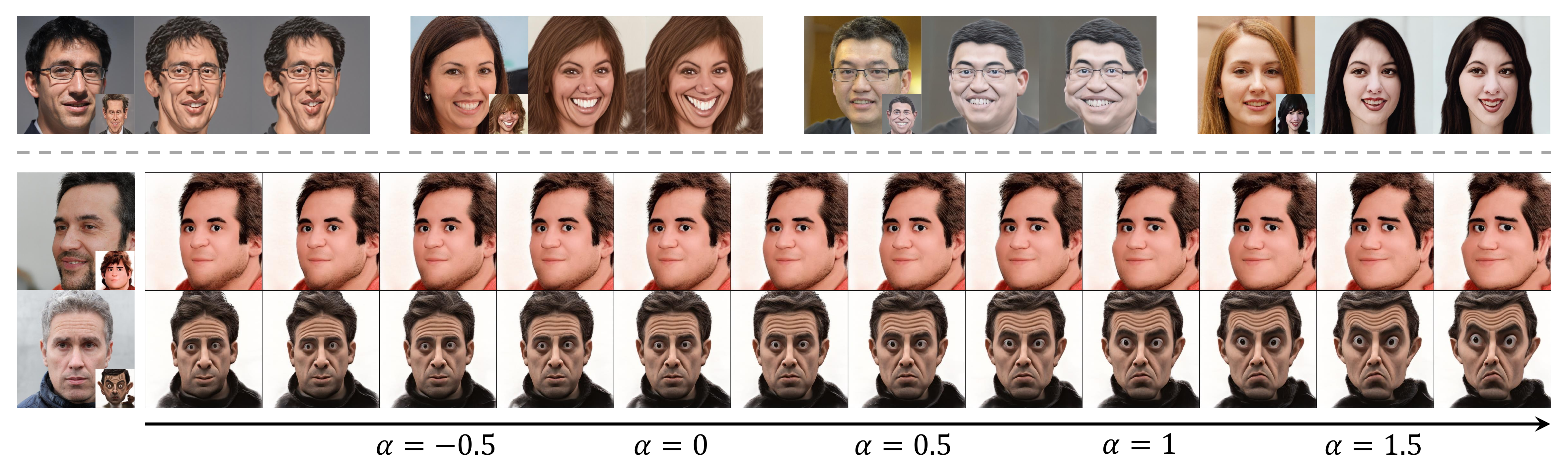}
        \caption{
    \textbf{ Controllable face deformation.} The STN blocks are functioned as plug-ins. We linearly interpolate the warping field of TPS-STNs to control the deformation degree of stylized faces. The triplets in the top row show the real faces, the stylized faces without and with STN-deformation, respectively. At the bottom, we showcase two stylized faces the deformation control of different degrees.
    }
    \label{fig:control}
    \vspace*{-2mm}
\end{figure*}

\begin{figure}[t!]
    \centering
    \includegraphics[width=1.\linewidth]{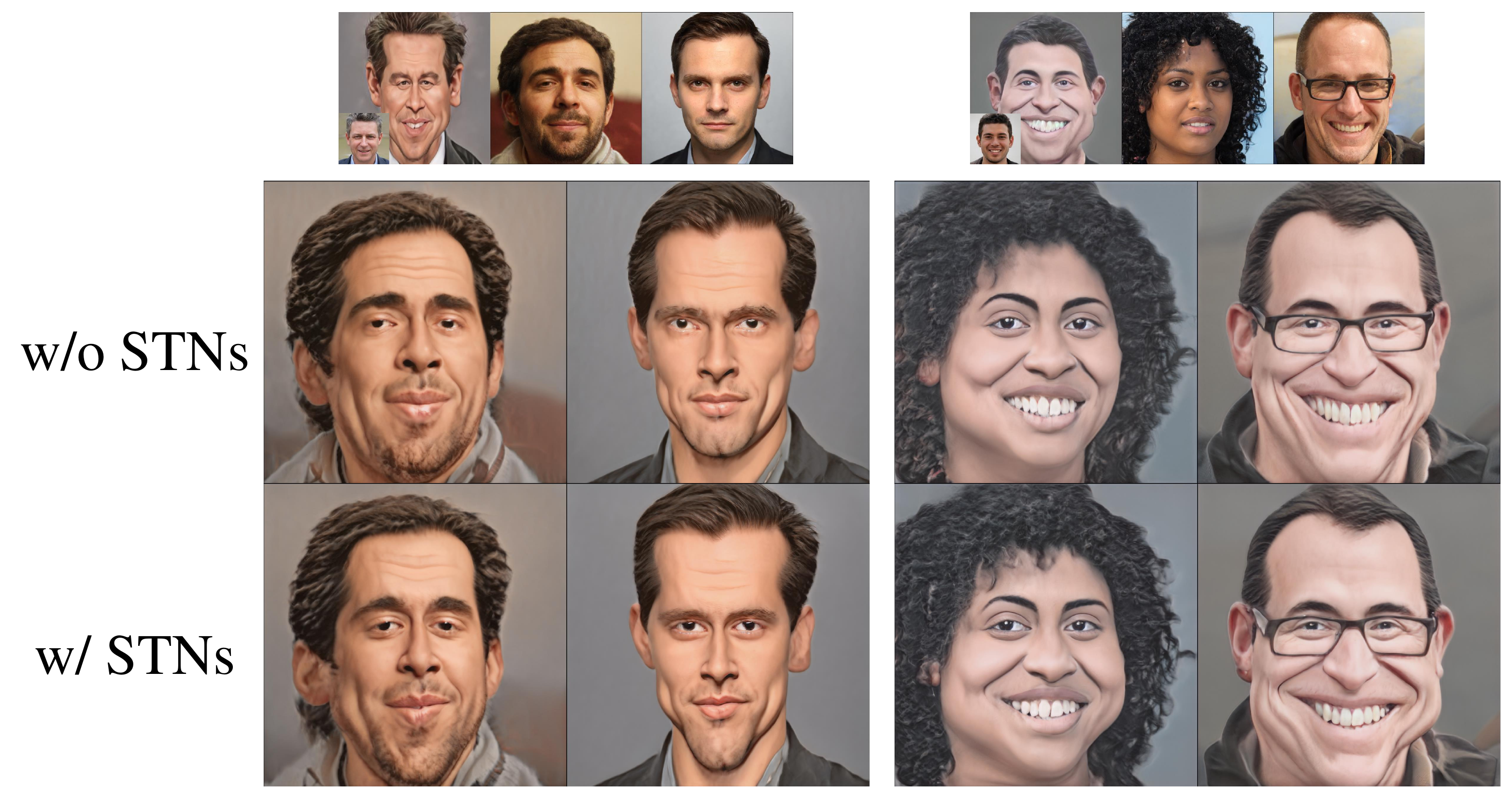}
        \caption{
    \textbf{ Ablation of the STNs appended in generator.} We fine-tune the generator with and without STN blocks. Some key exaggerated components, such as the eye contraction (left) and the cheek deformation (right), are missed when removing STN blocks.
    }
    \label{fig:abl_stn}
    \vspace*{-2mm}
\end{figure}

\begin{figure}[ht!]
    \centering
    \includegraphics[width=1.\linewidth]{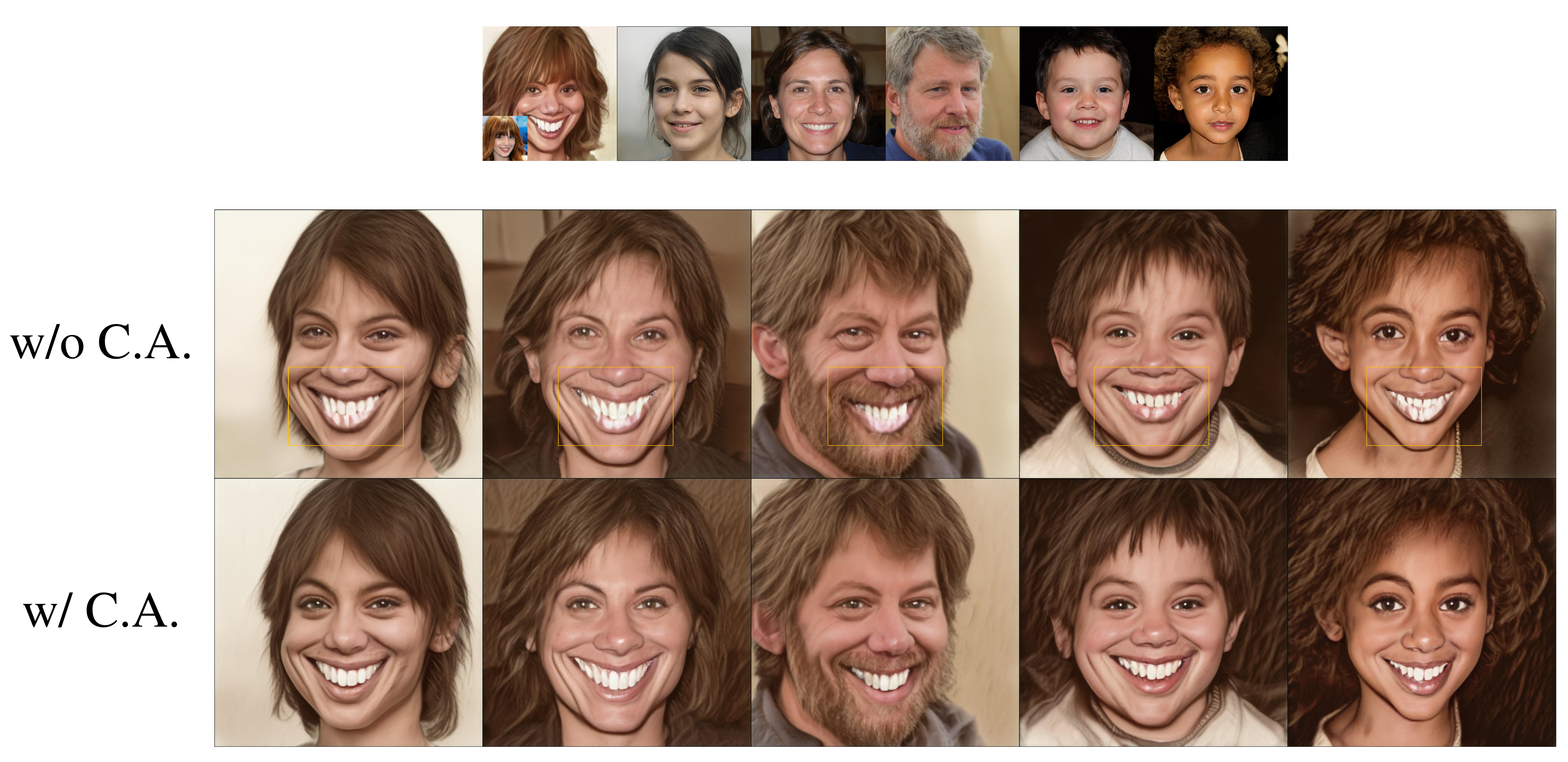}
        \caption{
    \textbf{ Ablation of color alignment.} Color alignment (C.A.) plays an important role in ensuring correct semantic matching when color gaps exist in the paired reference. Note how the lower lips (highlighted) are influenced when there is no color alignment.
    }
    \label{fig:abl_align}
    \vspace*{-2mm}
\end{figure}

\noindent\textbf{Effect of STN blocks.} 
As shown in Fig.~\ref{fig:abl_stn}, the generator with STN blocks brings a more accurate deformation effect regarding the desired exaggerations shown in the style reference, thus improving the stylization quality, as exaggeration is a key component in artistic styles.

\noindent\textbf{Effect of color alignment.} 
Fig.~\ref{fig:abl_align} verifies the effect of color alignment.
When a huge color gap exists in the paired reference, color alignment ensures a consistent DINO semantic representation, alleviating structural mismatching. 

\subsection{Facial deformation control}
\label{sec:control}
Since the STNs are plug-in blocks that warp the intermediate features of StyleGAN, we can use them to control the feature warping degree by interpolation. Specifically, we interpolate the warping field of TPS-STNs by a weight $\alpha$:
\begin{equation}
    \mathbf{F}(\alpha)=(1-\alpha)*\mathbf{F}_0+\alpha*\mathbf{F},
\end{equation}
where $\mathbf{F}_0$ is the original warping field without any warping. Setting $\alpha$=1 or $\alpha$=0, respectively, leads to results with or without the warping deformations. Fig.~\ref{fig:control} shows examples of such controllable face deformation in different styles.


\section{Conclusion}
We have presented a deformable face stylization framework using only a single paired reference. By fine-tuning a deformation-aware generator under DINO semantic guidance, we can stylize facial images with high-quality appearance transfer and convincing structure deformation. Qualitative and quantitative comparisons demonstrate that our method surpasses the existing one-shot methods. 

\paragraph{Acknowledgements}
This work was supported in parts by NSFC (U21B2023, U2001206, 62161146005), GD Natural Science Foundation (2022A1515010221), DEGP Innovation Team Program (2022KCXTD025), Shenzhen Science and Technology Program (KQTD20210811090044003, RCJC20200714114435012), Guangdong Laboratory of Artificial Intelligence and Digital Economy (SZ) and Scientific Development Funds of Shenzhen University. 

{\small
\bibliographystyle{ieee_fullname}
\bibliography{egbib}
}

\end{document}